# Greedy Ants Colony Optimization Strategy for Solving the Curriculum Based University Course Timetabling Problem

**Kenekayoro Patrick[1*] and Zipamone Godswill[1]**

[1]*Department of Mathematics and Computer Science, Niger Delta University, Amassoma, Bayelsa State, Nigeria.*

*Authors' contributions*

This work was carried out in collaboration between both authors. Author KP designed the study, wrote the protocol and supervised the work. Authors KP and ZG carried out all laboratories work and performed the statistical analysis. Author KP managed the analyses of the study. Author KP wrote the first draft of the manuscript. Author ZG managed the literature searches and edited the manuscript. Both authors read and approved the final manuscript.

___

## Abstract

Timetabling is a problem faced in all higher education institutions. The International Timetabling Competition (ITC) has published a dataset that can be used to test the quality of methods used to solve this problem. A number of meta-heuristic approaches have obtained good results when tested on the ITC dataset, however few have used the ant colony optimization technique, particularly on the ITC 2007 curriculum based university course timetabling problem. This study describes an ant system that solves the curriculum based university course timetabling problem and the quality of the algorithm is tested on the ITC 2007 dataset. The ant system was able to find feasible solutions in all instances of the dataset and close to optimal solutions in some instances. The ant system performs better than some published approaches, however results obtained are not as good as those obtained by the best published approaches. This study may be used as a benchmark for ant based algorithms that solve the curriculum based university course timetabling problem.

___

*\*Corresponding author: E-mail: patrick.kenekayoro@outlook.com;*

*Keywords: Meta-heuristics; ants colony optimization; university course timetabling; greedy algorithm.*

# 1 Introduction

The University Course Timetabling Problem (UCTP) is concerned with assigning venues and timeslots to courses whilst meeting several required or desirable constraints. There have been a number of approaches to solving this problem, some of which divide the problem into two stages, the construction stage that finds a solution that meets all required constraints and the improvement stage that improves the solution found in the construction stage by minimizing desirable constraints.

Some of the two stage approaches have solved the UCTP with local search methods like simulated annealing [1], tabu search [2] and hill climbing [3] to improve a feasible timetable that was previously constructed with heuristics such as the least saturation degree first heuristics [4]. The least saturation degree first heuristic sequentially assigns to the timetable solution harder courses (those courses that can be assigned to fewer room period combinations) before assigning less hard courses. If a feasible solution is not found in the first instance, the solution is repaired by applying neighborhood functions like swapping the time slots or venues of two events recursively until it is transformed to a feasible solution.

Evolutionary based local search techniques like ants colony [5], genetic [6] and swarm intelligence [7] algorithms can be used to efficiently find a feasible timetable solution that would not need to be repaired, and can produce different solutions for a given problem as feasible solutions are not constructed in a deterministic fashion. Evolutionary strategies can also select sequence or combination of moves adaptively which in turn may improve the quality of the resulting timetable [8,9].

This paper describes a greedy ants' colony optimization strategy for solving the university course timetabling problem. Ant systems have previously been used for timetable construction [5,10,11]. This study attempts to construct a feasible timetable in the first phase and then improves the desirable constraints on the found feasible timetable in the second phase, which is different from the ant colony optimization methods like [5,10,11] that optimize both required and desirable constraints in a single phase.

Subsequent sections in this paper formally describes the UCTP, the ants' system for construction and improvement stages and then the performance of the ants' colony algorithm is tested on ITC 2007 curriculum based benchmark dataset and compared with other published results. This dataset has been used as benchmark for testing the quality of algorithms in a number of researches, but no published result on the quality of solutions that an ants' system can find was identified for this dataset during the literature review. Nothegger et al.'s [11] ants' system solved the post enrolment university course timetabling problem.

# 2 University Course Timetabling Problem (UCTP)

The UCTP has been defined as the construction of a timetable where *"a number of courses must be placed within a given timetable (assigned venues and periods), so that the timetable is feasible (may actually be carried out), and a number of additional preferences to be satisfied is maximized"* [12]. A feasible timetable solution is met when all required constraints are met. Required constraints are hard constraints while desirable requirements (additional preferences) are soft constraints. A violation of a soft constraint does not result in an infeasible timetable solution. The goal is to minimize the soft constraint violations. The majority of researches have always found a feasible solution, sometimes by a simple least saturation degree first heuristic [2,13], thus the majority studies have focused on minimizing the soft constraints.

The ITC 2007 dataset is used for evaluating the algorithm, thus hard and soft constraints are the same as the constraints used in the Track 2 (UD2) of the competition as well as in a number of studies that have used this dataset [2,7,13–17]. These constraints as listed in [18] are as follows:

## 2.1 Hard constraints

HC1. All courses should be assigned the required number of weekly lectures.
HC2. A lecturer should not be assigned to teach more than one course at a given period.
HC3. Students should not be scheduled to attend more than one lecture at a given period.
HC4. Not more than one course can be assigned to a room at a given period.
HC5. Courses should be assigned to periods when the lecturer is available.

## 2.2 Soft constraints

SC1. The number of students attending a lecture in a particular room should be less or equal to the capacity of the room. Each student that will not be allocated a seat if a lecture is scheduled in a particular room counts as one violation.
SC2. The number of lecture days for each course should be greater or equal to the minimum working days for that course. For each course, the number of days less than the minimum working days is the number of violations.
SC3. Students' lectures should not be isolated, that is; lectures that belong to each curriculum in a given day should be in consecutive periods. Each lecture belonging to curriculum that is isolated counts as one violation.
SC4. All lectures of a course should be scheduled in the same room. Each additional room different from the first room given to a course counts as one violation.

Any hard constraint violation results in an infeasible timetable. The quality of a feasible timetable which depends on the number of soft constraint violations is evaluated by:

$$quality = Count_{SC1} * 1 + Count_{SC2} * 5 + Count_{SC3} * 2 + Count_{SC4} * 1$$

Lu and Hao [2] defined the detailed mathematical model of the curriculum based UCTP with the hard and soft constraint violations for the track 2 of the ITC 2007 dataset. The mathematical model for this study is identical to the model described in Lu and Hao [2].

# 3 Algorithm

This paper uses ants' colony optimization for both construction and improvement phases, which is different from other ant systems that finds a solution in a single phase.

## 3.1 Construction phase

A number of studies have used heuristic techniques like the least saturation degree first and largest degree first heuristics [1,3,9,19,20], which constructs a feasible solution by assigning the most difficult events to a room/time slot before the less difficult events. Difficult events are those events that have fewer available rooms/time slots that they can be assigned to. In some cases, a feasible solution is not be found by the construction heuristics in a single run [15], the solution can be repaired by neighbourhood search [9].

Geiger [15] adaptively updated the order in which events are to be scheduled by identifying those events that are particularly difficult to schedule in the previous run and then giving them priority in subsequent run, which can be seen as the working principle of the ants colony optimization algorithm.

Algorithm 1 and Algorithm 2 describes the ant system that finds a feasible solution.

### 3.1.1 Representation

Problems have to be represented as a graph before a solution can be found using ants' colony optimization. Socha, Knowles and Sampels [5] represented the UCTP as a graph of virtual timeslots. Their representation does not take room assignments into account but uses a deterministic network flow algorithm for room assignments. Djamarus and Ku-Mahamud [10] modelled the UCTP as a bipartite graph of course vertices, lecturer vertices, period vertices and room vertices which may result in a large pheromone trail matrix. Nothegger et al. [11] represented the UCTP with course, room and period vertices, but stored the pheromone trail information in two matrices, $PeriodTrail_{i,j}$; the trail for event i in time slot j, and $RoomTrail_{i,k}$; the trail for event i in room k. The authors noted that is not as expressive as representing the trail information as $Trail_{i,j,k}$; the trail of event i in timeslot j and in period k, but it sufficiently approximates $Trail_{i,j,k}$ and is computationally less expensive. This study uses the more expressive representation, accepting the trade-off of fewer computations per unit time. Room and time slots were combined to a single vertex and edges connect a Course $C_i$ to a RoomPeriod $R_jP_k$ if scheduling the course in that room and period does not violate any unavailable period constraint (HC5) and the difference in room capacity and course size is not more than one; thus minimizing SC1. This representation ensures that HC5 cannot be violated and SC1 is minimized in any timetable solution generated. The illustration of the graph representation is shown in Fig. 1.

### 3.1.2 Initialization

The pheromone trail matrix is initialized using Min Max Ant System Algorithm [5]. The value (edge weight) in the trail matrix is initialized to $T_{max}$ if an edge exists between the course vertex and RoomPeriod vertex (assigning a course that period will not violate HC5 and the difference between the room capacity and course size is not more than one), else the value on the trail matrix is initialized to zero.

### 3.1.3 Graph walk

An ant starts at a random Course vertex, and then walks to a possible RoomPeriod vertex with a probability dependent on the quantity of pheromone on the edge, adding visited course, room and period triples to an initially empty partial timetable solution, until all events have been scheduled. The probability that an ant will follow an edge is computed using Equation (1). Possible RoomPeriod vertices are those vertices that will not violate any hard constraint when the course, room and period event triple is added to the partial timetable solution.

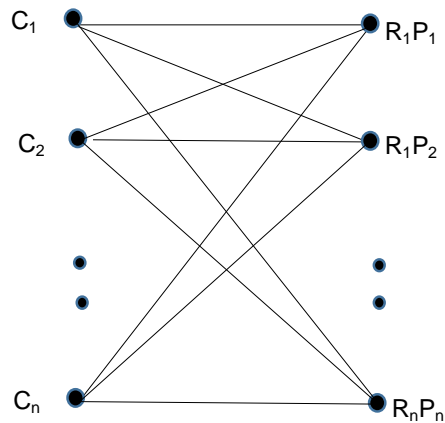

**Fig. 1. Graph representation of the university course timetabling problem**

If no such RoomPeriod vertex exists, a RoomPeriod vertex that will violate a hard constraint is walked to with a probability based on the pheromone trail on the edge and the number of hard constraints that will be violated (heuristic) when the course, room period event triple is added to the partial solution. The edge weight on the trail matrix is also not updated during trail update.

$$Probability_{i,j} = \frac{Trail_{i,j}^{\alpha} + Heuristic_{i,j}^{\beta}}{\sum_{k=1}^{number\ of\ possible}\left(Trail_{i,j}^{\alpha} + Heuristic_{i,j}^{\beta}\right)} \quad (1)$$

Alpha, α is the trail weight, a high value increases the sensitivity of ants to pheromone trail. Beta, β is the heuristic weight, a high value increases the sensitivity of ants to heuristic information. Heuristic is computed as number of hard constraints (5) minus number of additional hard constraint violations if Course i and RoomPeriod j event tuple is added to the partial timetable solution. After several tests, Alpha = 2 and Beta = 8 is used for the results shown in subsequent sections.

### 3.1.4 Trail update

It is suggested that not all ants should update the pheromone trail [21], thus the elitist strategy where only the ant with the best solution in a cycle (CBest) updates the pheromone trail [22] is used. After all ants in a cycle have walked the UCTP graph and found a solution, the best solution (smallest number of hard constraints violations) is used to update the trail matrix. All edges that the ant followed to find its solution, except those edges that caused hard constraint violations are updated using the formula in Equation (4). The algorithm used here only rewards positively, different from [10] that positively or negatively updates ant trails. In tests, computing reward with only positive updates Equation (4) found feasible solutions faster than computing rewards with Equation (3) that is described in [22]. The values on the pheromone trail matrix are reduced through the evaporation formula in Equation (5) in order to avoid convergence to a local minimum. The rate of evaporation is controlled by rho; ρ, which can be between 0 and 1. When ρ is high, the rate of evaporation is high and thus convergence is slow. When ρ is small, the rate of evaporation is slow, convergence is fast and thus reduces the search space ants explore.

$$Reward = \begin{cases} \frac{1}{1+CBest-GBest}, & CBest > GBest \\ 1, & CBest < GBest \end{cases} \quad (2)$$

$$Reward = \frac{1}{1+GBest-CBest} \quad (3)$$

$$Trail_{1,j} = Trail_{i,j} + Reward \quad (4)$$

$$Trail_{1,j} = Trail_{i\,j} * (1-\rho) \quad (5)$$

| | **Algorithm 1 construction graph walk** |
|---|---|
| **1** | **AntWalk(alpha, beta, trail)** |
| 2 | While not visited all courses |
| 3 | C = random course not already visited |
| 4 | While not assigned all weekly lectures |
| 5 | Possible = List of feasible room period objects the course can be assigned to |
| 6 | RP = a RoomPeriod from possible based on probability in Equation **(1)** |
| 7 | Create Event from tuple (C,R,P) |

| | **Algorithm 2 ants colony optimization** |
|---|---|
| 1 | **AntColonyOptimization (numberCycles, numberAnts)** |
| 2 | Input: |
| 3 | alpha = 2 |
| 4 | beta = 8 |
| 5 | rho = 0.05 |
| 6 | tmin = 0.01 |
| 7 | tmax = 10 |
| 8 | GBest = ∞ |
| 9 | Initialize Pheromone trail matrix |
| 10 | for i=1 to numberCycles |
| 11 | for j=1 to numberAnts |
| 12 | solution[j] = AntWalk(alpha, beta, trail) |
| 13 | CBest = $\boldsymbol{MinimumConstraintViolation_{j=1}^{NumberOfAnts}}$ |
| 14 | if CBest < GBest |
| 15 | GBest = CBest |
| 16 | if GBest = 0 |
| 17 | return GBest Timetable Solution |
| 18 | Update pheromone trail matrix |
| 19 | Pheromone trail evaporation |
| 20 | Set edge weights to tmax if edge weight > tmax |
| 21 | Set edge weights to tmin if edge weight < tmin |

### 3.2 Improvement phase

The improvement phase is key feature of the method used in this study. Local search algorithms like hill climbing, great delunge, simulated annealing and tabu search [3,9,23,24] have been used for the improvement phase of exam or course timetabling problem, but few studies have used ant systems. Nothegger et al. [11] used a local search algorithm that does not escape the local optima, however the main algorithm used in their study minimized both hard and soft constraints, unlike [3,9,23,24] that does not take soft constraint violations into account in the construction phase.

The ant system used in the improvement phase of this study improves the quality of a feasible timetable with a greedy heuristics. Only improving moves are accepted which makes this essentially a hill climbing approach, however pheromone trails are used to break ties.

#### 3.2.1 Neighbourhood structures

In the improvement phase, neighbourhood structures are used to improve the quality (reduce the number of soft constraint violations) in an already feasible schedule. The choice of neighbourhood structures is important for local search algorithms as they are believed to be among key features that determine the quality of the local search algorithm [2].

The simple swap neighbourhood structure described in Lü and Hao [2] is also used in this study. The courses of two course room period triples are exchanged. For example, the result of swapping $C_1RP_1$ with $C_2RP_2$ results in $C_2RP_1$ and $C_1RP_2$. A move occurs when the course in one of the two events (Course Room Period triple) is empty.

#### 3.2.2 Representation

The improvement problem is also represented as a graph whose adjacent matrix is shown in Table 1. The move $[i,j]$ creates a new schedule where the course at $CRP_i$ is swapped with the course at $CRP_j$. The maximum value of $i$ is the number of events in the schedule, while the maximum number of $j$ is the number of events + number of room period tuples. This representation sometimes results in a large matrix which could be computationally expensive.

**Table 1. Adjacent matrix representation of ants system for improvement phase**

| $CRP_1$ | $CRP_1$ | $CRP_2$ | ... | $CRP_n$ | $EmptyRP_1$ | $EmptyRP_2$ | ... | $EmptyRP_n$ |
|---|---|---|---|---|---|---|---|---|
| $CRP_2$ | $CRP_1$ | $CRP_2$ | ... | $CRP_n$ | $EmptyRP_1$ | $EmptyRP_2$ | ... | $EmptyRP_n$ |
| . | . | . | ... | . | . | . | ... | . |
| . | . | . | ... | . | . | . | ... | . |
| . | . | . | ... | . | . | . | ... | . |
| $CRP_n$ | $CRP_1$ | $CRP_2$ | ... | $CRP_n$ | $EmptyRP_1$ | $EmptyRP_2$ | ... | $EmptyRP_n$ |

**3.2.3 Initialization**

All elements on the trail matrix are initialized to $T_{max}$, because depending on previous swaps, subsequent swap moves may or may not result in a feasible schedule.

Graph walk and trail update are the same as described in the construction phase. The Ant walk algorithm is shown in Algorithm 3.

| | **Algorithm 3 improvement graph walk** |
|---|---|
| 1 | **AntWalk(alpha, beta, trail)** |
| 2 | While not visited all events |
| 3 | $CRT_1$ = random event not already visited |
| 4 | Possible=List of CRT swap with lowest soft constraint violations |
| 5 | $CRT_2$=a CRT from possible based on probability in Equation (**1**) |
| 6 | Apply neighbourhood structure ($CRT_1$, $CRT_2$) |

Combining multiple algorithms have been shown to result in good solutions to optimization problems [3,13], when simulated annealing, hill climbing and great deluge algorithms were combined in a round robin sequence to solve the university course timetabling problem. In this study, only one algorithm (ants' colony optimization) is used, however, the ants' colony algorithm is run multiple times. The input of the next run is the output of the previous run.

# 4 Results and Discussion

Table 2 reports the results of running the greedy ants' colony optimization algorithm described in earlier sections. After tests, it was determined that 8 ants were suitable for the algorithm.

In the construction phase, the algorithm runs until a feasible solution is found, which is instant in some instances, except difficult instances like comp05 where over 100 cycles is needed to find a feasible solution. A feasible solution was always found in all instances in less than 350 seconds, which is less than the time out condition for in the ITC-2007 competition rules.

In the improvement phase, the algorithm was run until 10 consecutive solutions that are not improvements on the current global best is reached. The representation used to improve soft constraint violations in this study is expensive, because of the large number of evaluations that are computed in each iteration. An alternative is using only ant trails to direct ants, but that will slow convergence even though it may produce better results. Thus, a more efficient representation is needed in the improvement stage.

Each ant walk in independent, thus ant walks in a cycle is implemented to run concurrently with different threads. This increases the number of computations in a given time interval, thus solutions are found faster.

In terms of the quality of the results, the proposed algorithm does not achieve results better than the best published algorithms, however it outperforms some algorithms [7,14]. In 17 out of the 21 instances, the proposed methods' soft constraint violations is less than the mean of all algorithms compared. Lü and Hao [2] have emphasized on the importance of neighbourhood structures in local search algorithms, so a more complex neighbourhood function may be used to achieve better results than the result stated in Table 2.

The algorithm is implemented in C++, compiled with GCC 4.8.4 on a 64 bit Ubuntu 14.04 machine (Intel® Core™ i7-3612QM CPU @ 2.10GHz × 8 and 8.0GB memory).

**Table 2. Comparison of the Proposed Method (PM) and other published solutions including the best known solution as published in [2]**

| Instance | PM  | A1   | A2  | A3  | A4  | A5  | A6  | A7  | BK  | Mean   |
|----------|-----|------|-----|-----|-----|-----|-----|-----|-----|--------|
| Comp01   | 10  | 322  | 5   | -   | 5   | 9   | 5   | 5   | 5   | 51.57  |
| Comp02   | 176 | 732  | 108 | 312 | 43  | 103 | 29  | 43  | 29  | 193.25 |
| Comp03   | 222 | 665  | 115 | 292 | 72  | 101 | 66  | 77  | 66  | 201.25 |
| Comp04   | 100 | 577  | 67  | 193 | 35  | 55  | 35  | 38  | 35  | 137.50 |
| Comp05   | 606 | 1297 | 408 | -   | 298 | 370 | 292 | 311 | 292 | 511.71 |
| Comp06   | 178 | 879  | 94  | 336 | 41  | 112 | 37  | 44  | 37  | 215.13 |
| Comp07   | 123 | 930  | 56  | 324 | 14  | 97  | 13  | 19  | 7   | 197.00 |
| Comp08   | 112 | 645  | 75  | 218 | 39  | 72  | 39  | 44  | 38  | 155.50 |
| Comp09   | 172 | 685  | 153 | 302 | 103 | 132 | 96  | 108 | 96  | 218.87 |
| Comp10   | 125 | 816  | 66  | 274 | 16  | 74  | 10  | 13  | 7   | 174.25 |
| Comp11   | 1   | 179  | 0   | 293 | 0   | 1   | 0   | 0   | 0   | 59.25  |
| Comp12   | 622 | 1398 | 430 | -   | 331 | 393 | 310 | 339 | 310 | 546.14 |
| Comp13   | 136 | 694  | 101 | -   | 66  | 97  | 59  | 69  | 59  | 174.57 |
| Comp14   | 141 | 702  | 88  | 236 | 53  | 87  | 51  | 60  | 51  | 177.25 |
| Comp15   | 189 | 665  | 128 | 284 | 84  | 119 | 68  | 76  | 68  | 201.63 |
| Comp16   | 155 | 827  | 81  | 281 | 34  | 84  | 23  | 48  | 23  | 191.63 |
| Comp17   | 148 | 830  | 124 | 331 | 83  | 152 | 69  | 91  | 69  | 228.50 |
| Comp18   | 132 | 510  | 116 | 196 | 83  | 110 | 65  | 84  | 75  | 162.00 |
| Comp19   | 156 | 608  | 107 | 304 | 62  | 111 | 57  | 71  | 57  | 184.50 |
| Comp20   | 147 | 950  | 88  | 372 | 27  | 144 | 22  | 42  | 17  | 224.00 |
| Comp21   | 246 | 835  | 174 | -   | 103 | 169 | 93  | 103 | 89  | 246.14 |

*PM - Proposed Greedy Ant System, A1 [14], A2 [15], A3 [7], A4 [16], A5 [17], A6 [2], A7 [13], BK - Best Known results as published in [2]*

## 5 Conclusions

This study has used ants' colony optimization to solve the university course timetabling problem. Previously, a number of heuristics techniques have been used to solve this problem however few have used the ants system, particularly for the curriculum based timetabling problem in the ITC-2007 dataset.

Although the results obtained by the proposed ants' system is not as good as state of art techniques, it outperforms a number of published results. Also, as this is among the few that have used the ants colony algorithm in this dataset, the results published here can be used as benchmarks for other ant systems using this dataset.

The ants system in the improvement phase is computationally expensive, thus future research well aim to address this, as well as improve on the results so that it is more competitive with state of art techniques.

## Competing Interests

Authors have declared that no competing interests exist.

## References

[1]  Tuga M, Berretta R, Mendes A. A hybrid simulated annealing with kempe chain neighborhood for the University Timetabling Problem. Computer and Information Science, ICIC 2007 6th IEEE/ACIS International Conference. 2007;400–405.

[2]  Lü Z, Hao JK. Adaptive tabu search for course timetabling. European Journal of Operational Research [Internet] Elsevier B.V. 2010;200(1):235–244. [cited 2014 Oct 27].
Available: http://linkinghub.elsevier.com/retrieve/pii/S0377221708010394

[3]  Shaker K, Abdullah S. Controlling multi algorithms using round robin for university course timetabling problem. Database Theory and Application, Bio-Science and Bio-Technology. 2010; 47–55.

[4]  Burke EK, Mccollum B, Meisels A. Discrete optimization a graph-based hyper-heuristic for educational timetabling problems. 2007;176:177–192.

[5]  Socha K, Knowles J, Sampels M. A MAX - MIN ant system for the university course timetabling problem. Lecture notes in computer science: Proceedings of the 3rd international workshop on ant algorithms. Springer Berlin. 2002;1–13.

[6]  Rahoual M, Saad R, Timetabling S, Genetic H. Solving Timetabling problems by hybridizing genetic algorithms and taboo search. Proceedings of the 6th International Conference on Pratice and Theory of Automated Timetabling Brno, Czech Republic. 2006;467–472.

[7]  Bolaji AL, Khader AT, Al-betar MA, Awadallah M. Artificial Bee colony algorithm for curriculum-based course timetabling problem. The 5th International Conference on Information Technology; 2011.

[8]  Ochoa G, Qu R, Burke EK. Analyzing the Landscape of a Graph Based Hyper-heuristic for Timetabling Problems; 2009.

[9]  Abdullah S, Shaker K, Mccollum B, Mcmullan P. Dual sequence simulated annealing with round-robin approach for university course timetabling. Evolutionary Computation in Combinatorial Optimization Springer-Verlag. 2010;1–10.

[10] Djamarus D, Ku-Mahamud KR. Heuristic Factors in Ant System Algorithm for Course Timetabling Problem. 2009 Ninth International Conference on Intelligent Systems Design and Applications [Internet] Ieee. 2009;232–236. [cited 2014 Oct 27]
Available: http://ieeexplore.ieee.org/lpdocs/epic03/wrapper.htm?arnumber=5364796

[11] Nothegger C, Mayer A, Chwatal A, Raidl GR. Solving the post enrolment course timetabling problem by ant colony optimization. 2012;325–339.

[12] Socha K, Sampels M, Manfrin M. Ant algorithms for the university course timetabling problem with regard to the state-of-the-art. Applications of evolutionary computing Springer Berlin Heidelberg. 2003;335–345.

[13] Abdullah S, Shaker K, Shaker H. Investigating a round robin strategy over multi algorithms in optimising the quality of university course timetables. 2011;6(6):1452–1462.

[14] Wahid J. Harmony search algorithm for curriculum-based course timetabling problem. arXiv preprint arXiv:14015156; 2014.

[15] Geiger MJ. An application of the threshold accepting metaheuristic for curriculum based course timetabling. Proceedings of the 7th International Conference on Pratice and Theory of Automated Timetabling; 2008.

[16] Muller T. ITC2007 solver description: A hybrid approach. Annals of Operations Research. 2009;172(1):429–446.

[17] Clark M, Henz M, Love B. QuikFix a repair-based timetable solver. Proceedings of the 7th International Conference on Pratice and Theory of Automated Timetabling; 2008.

[18] De Cesco F, Di Gaspero L, Schaerf A. Benchmarking curriculum-based course timetabling: formulations, data formats, instances, validation, and results. Proceedings of the 7th International Conference on the Practice and Theory of Automated Timetabling, PATAT. 2008;1–11.

[19] Sabar NR, Ayob M, Kendall G, Qu R. Discrete optimization a honey-bee mating optimization algorithm for educational timetabling problems. European Journal of Operational Research [Internet] Elsevier B.V. 2012;216(3):533–543.
Available: http://dx.doi.org/10.1016/j.ejor.2011.08.006

[20] Burke EK, Mccollum B, Meisels A, Petrovic S, Qu R. A Graph-based hyper-heuristic for educational timetabling problems. European Journal of Operational Research. 2007;176(1):177–192.

[21] Dorigo M, Maniezzo V, Colorni A. Ant system: Optimization by a colony of cooperating agents. Systems, Man, and Cybernetics, Part B: Cybernetics, IEEE Transactions. 1996;26(1):29–41.

[22] Solnon C, Bridge ID. An ant colony optimization meta-heuristic for subset selection problems. System Engineering using Particle Swarm Optimization, Nova Science. 2006;7–29.

[23] Sabar NR, Ayob M, Kendall G. Tabu exponential monte-carlo with counter heuristic for examination timetabling. Computational Intelligence in Scheduling, 2009 CI-Sched'09 IEEE Symposium on IEEE. 2009;90–94.

[24] Zhaohui F, Lim A. Heuristics for the exam scheduling problem. Proceedings 12th IEEE Internationals Conference on Tools with Artificial Intelligence ICTAI 2000 [Internet] IEEE Comput. Soc. 2000;172–175.
Available: http://ieeexplore.ieee.org/lpdocs/epic03/wrapper.htm?arnumber=889864